\documentclass[letterpaper, 10 pt, conference]{ieeeconf}  

\IEEEoverridecommandlockouts                              

\usepackage{times}
\usepackage{epsfig}
\usepackage{graphicx}
\usepackage{amsmath}
\usepackage{amssymb}
\usepackage{listings}
\usepackage{tabu}

\usepackage{hyperref}
\hypersetup{
    colorlinks=true,
    linkcolor=blue,
    filecolor=magenta,      
    urlcolor=cyan,
}
 
\title{\LARGE \bf
Fast Disparity Estimation using Dense Networks*
}

\author{Rowel Atienza$^{1}$
\thanks{*This work was funded by CHED-PCARI Project IIID-2016-005}
\thanks{$^{1}$Electrical and Electronics Engineering Institute, University of the Philippines
        {\tt\small rowel@eee.upd.edu.ph}}%
}

\begin{document}

\maketitle
\thispagestyle{empty}
\pagestyle{empty}

\begin{abstract}

   Disparity estimation is a difficult problem in stereo vision because the correspondence technique fails in images with textureless and repetitive regions. Recent body of work using deep convolutional neural networks (CNN) overcomes this problem with semantics. Most CNN implementations use an autoencoder method; stereo images are encoded, merged and finally decoded to predict the disparity map. In this paper, we present a CNN implementation inspired by dense networks to reduce the number of parameters. Furthermore, our approach takes into account semantic reasoning in disparity estimation. Our proposed network, called \textit{DenseMapNet}, is compact, fast and can be trained end-to-end. \textit{DenseMapNet} requires 290k parameters only and runs at 30Hz or faster on color stereo images in full resolution. Experimental results show that \textit{DenseMapNet} accuracy is comparable with other significantly bigger CNN-based methods. 

\end{abstract}

\section{INTRODUCTION}

Stereo vision contains rich sensory data that can be processed into more meaningful and useful information. With stereo images, a machine can estimate depth, optic flow, ego-motion, target object position, orientation, motion and 3D structure. Although classical computer vision offers hand engineered solutions, the techniques have oftentimes many limitations. For example, in disparity estimation, algorithms tend to break when given patches of textureless regions and repetitive areas like sky, road, floor, ceiling, and wall. The limitations may be primarily attributed to hand crafted cost functions that are unable to encompass multitude of possible observable scenarios. 

Furthermore, beyond measurements, classical techniques find it difficult to understand semantics. For example, the sky is known to be very far away whose depth could not be reliably estimated yet classical algorithms will still output estimates. If machines aim to achieve superhuman performance, they should understand semantics especially of the 3D environment.

In recent years, deep learning techniques have overcome the limitations of classical computer vision in areas of object detection, recognition, segmentation, depth estimation, optical flow, ego-motion and SLAM to name a few. For disparity, the use of CNN has significantly improved the accuracy, robustness and speed of measurement \cite{kendall2017end, mayer2016large, luo2016efficient, zbontar2015computing}.  In some cases, CNN is used in unsupervised depth estimation from monocular images \cite{godard2017unsupervised,zhou2017unsupervised}.

Current CNN-based implementations are designed either to mimic the classical correspondence technique at the pixel level or to predict the disparity image from feature maps of stereo images. Pixel level correspondence is unfavorable since it generally does not understand semantics of images being processed. It is just emulating the patch-based correspondence technique of classical computer vision using deep learning techniques. Using feature maps to predict disparity image is preferable since it takes into account the meaning of stereo images. Current feature maps-based approaches process stereo images into a latent representation using a CNN encoder. The latent representation is decoded by a transposed CNN to arrive with dense disparity estimates. Since stereo images have generally high resolution, the resulting autoencoder is deep and requires millions of parameters. An undesirable consequence of deep networks is gradient decay; weights and biases updates vanish as they propagate down to the shallow layers. Without residual network connections \cite{he2016deep} \cite{srivastava2015training}, the network is difficult to train.

\begin{figure}
\includegraphics[scale=0.28]{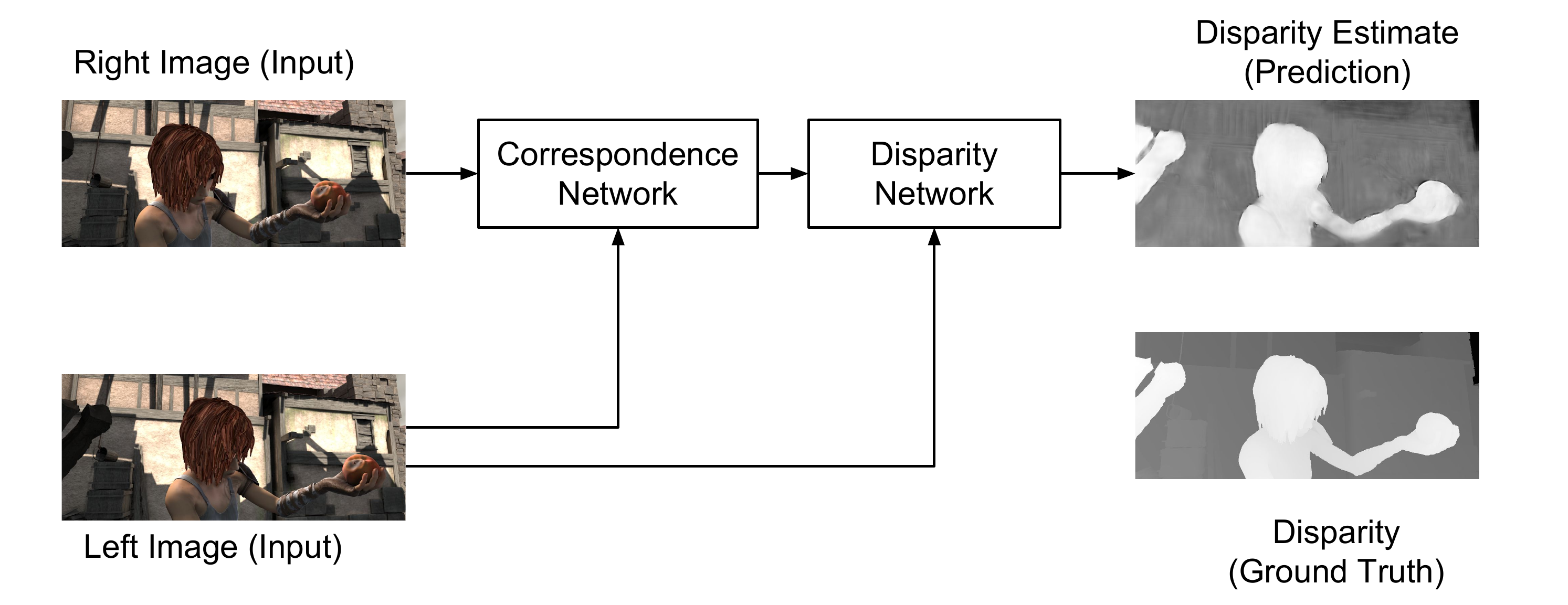}
\caption{\textit{DenseMapNet} is made of \textit{Correspondence Network} that finds correspondences between stereo images and \textit{Disparity Network} that applies the output of the \textit{Correspondence Network} to the reference image.}
\label{fig:densemapnetmodel}
\end{figure}

In this paper, we propose to use the technique of Dense Convolutional Networks (\textit{DenseNet}) \cite{huang2017densely} to address the vanishing gradient problem of deep networks. In \textit{DenseNet}, the input and output of each convolution feed the succeeding convolution. This prevents gradient decay since the loss function has immediate access to all layers. Furthermore, as shown in Figure~\ref{fig:densemapnetmodel}, we would like to use global reasoning in disparity estimation. We argue that disparity estimation can be generally decomposed into two networks. \textit{Correspondence Network} learns how to find correspondences given stereo images. \textit{Disparity Network} generates the final disparity map from the output of \textit{Correspondence Network} and a reference image.

Our proposed network, \textit{DenseMapNet}, is compact and requires 290k parameters only compared to 3.5M or more parameters in other similar CNN-based approaches. As a consequence of its small size, \textit{DenseMapNet} is fast. It can process color stereo images in full resolution at 30Hz or faster compared to the state-of-the-art at 16Hz. This is not to mention that we are running on a slower GPU, NVIDIA GTX 1080Ti, compared to the current state-of-the-art that used NVIDIA Titan X. The experimental results on benchmark datasets show that \textit{DenseMapNet} has accuracy comparable to other bigger CNN-based methods on both real and synthetic images. 

\section{RELATED WORK}

Given a pair of rectified images, the disparity of a pixel at $(x,y)$ on the left image is the offset $d$ of its location at $(x-d,y)$ on the right image. The depth $z$ can be computed as:

\begin{equation}\label{eq:1}
\mathit{z} = \frac{\mathit{fB}}{d}
\end{equation}

where $B$ is the stereo cameras baseline and $f$ is the camera focal length. Both constants can be measured through camera calibration. Depth measurement has applications in autonomous vehicle navigation, 3D reconstruction of observed objects, SLAM, robot motion planning, etc.

Scharstein and Szeliski \cite{scharstein2002taxonomy} discussed a taxonomy of algorithms in stereo correspondences. Generally, stereo matching uses one or more building blocks in the form of: 1) Matching computation, 2) Cost aggregation, 3) Disparity computation/optimization, and 4) Disparity refinement. In the classical computer vision, there has been numerous stereo correspondence algorithms. The local algorithms attempt to minimize a cost function between image patches such as in Sum of Absolute Differences (SAD), Sum of Squared Differences (SSD), and Normalized Cross Correlation (NCC) \cite{lewis1995fast}. The global algorithms aim to minimize a global function with smoothness assumptions to arrive at disparity estimates. Examples are Semi-Global Matching (SGM) \cite{hirschmuller2008stereo} and Markov Random Fields (MRF) \cite{zhang2007estimating} that minimize a global cost function and energy term.

Recently, deep learning techniques have overtaken benchmarks of disparity estimation accuracy on public datasets such as KITTI 2012 and 2015 \cite{Geiger2012CVPR, Menze2015CVPR,geiger2013vision} and Scene Flow \cite{mayer2016large}. The approaches can be roughly grouped into mimicking patch-based correspondences of classical method and image-based method where an autoencoder style model learns to map stereo images to disparity estimates.

In patch-based correspondences \cite{zbontar2015computing, luo2016efficient, chen2015deep}, the strategy is to compute the correspondence matching cost between the left image patch and the candidate right image patch. The feature map or vector of each patch is computed using CNN and sometimes with Multilayer Perceptron (MLP). The left and right feature vectors are combined to form a siamese network that predicts the best candidate match, hence the disparity, by minimizing the cross-entropy loss. Since the input stereo images are rectified, the search for a match is limited to the corresponding pixel row on the right image. The patch-based method differs slightly in the way the feature maps are combined. MC-CNN \cite{zbontar2015computing} concatenates the feature maps while Content-CNN \cite{luo2016efficient} and Chen et al. \cite{chen2015deep} compute the dot product to speed up the computation. The patch-based method can not be trained end-to-end. It still requires post-processing such as cost aggregation, semi-global matching, left-right consistency check, correlation, sub-pixel enhancement and interpolation to arrive at predicted disparity maps.  

The clear disadvantage of the patch-based method is it is just replacing the matching algorithm of the classical method with deep neural networks. The rest of the pipeline is still compute intensive because of the huge number of patches to consider even if GPU parallel processing is involved. Furthermore, there is little semantics that one can achieve at the pixel level. There are still post-processing steps involved preventing a complete end-to-end learning.

The image-based method uses an autoencoder style model that processes stereo images into dense disparity estimates, GC-Net \cite{kendall2017end} and DispNet \cite{mayer2016large}. The CNN encoder generates feature maps that feed the CNN decoder. To process stereo images simultaneously, a siamese network of CNN encoder with shared weights combines the feature maps. In the case of GC-Net, the feature maps are combined into a 3D cost volume which is then processed by a 3D convolution-deconvolution decoder. The model is trained by minimizing cross-entropy or MSE loss. Within the autoencoder, additional layers are used to increase the disparity estimate accuracy like the correlation network in DispNet and residual network in GC-Net. Both networks are deep with 26 layers for DispNet and 37 layers for GC-Net and up to 1024 feature maps per CNN. Both image-based methods use cropped images only as input to the encoder. This is understandable considering the KITTI datasets images are at least $1240\times{376}$ pixels in size \cite{geiger2013vision,Menze2015CVPR,Geiger2012CVPR} and MPI Sintel dataset images are $1024\times{436}$ pixels \cite{butler2012naturalistic}. Using cropped images as input reduces the size of the network. However, this could have a negative impact on the prediction of disparity near the boundary of the cropped images.  The output dense disparity map is of the same size for GC-Net or half the size as the cropped image for DispNet. Up sampling or super resolution is applied to get the full disparity map in case the predicted image is of lower dimensions.

It is evident that like other deep CNN, disparity estimation suffers the problem of vanishing gradient. Both GC-Net and DispNet address the problem by using sparse residual connection - occasionally connecting lower-layer feature maps to higher layers. DispNet also uses auxiliary loss functions to prevent gradient decay.  Recently, \textit{DenseNet} \cite{huang2017densely} introduced a new CNN architecture wherein all layers are connected to all previous inputs and outputs. \textit{DenseNet} argues that most of the time residual networks have residual layers with little contribution or effect in the final prediction. Since residual layers have their own set of parameters, the number of weights to train increases unnecessarily. In \textit{DenseNet} all previous layers inputs are combined with the current layer inputs and share the same set of parameters. In effect, \textit{DenseNet} uses significantly less number of parameters. 

Our proposed model, \textit{DenseMapNet}, as shown in  Figure~\ref{fig:densemapnetmodel} differs in few key areas. First, we utilize full resolution images on both input and output. There is no image cropping in the pipeline. Instead, we use max pooling and up sampling to manage the number of parameters and memory use of the network. Second, to resolve the problem of vanishing gradient and to reduce the number of parameters to train, our model has each layer connected to all previous layers similar to \textit{DenseNet}. Third, it is known that to generate disparity map we use one image as reference (e.g. left image) and increase the intensity of each pixel in proportion to its disparity (by finding its corresponding match on the right image using \textit{Correspondence Network}). We use this reasoning to guide our design and arrive at a two-network model - a \textit{Correspondence Network} and a \textit{Disparity Network}. Each layer has direct access to all previous output layers and stereo images. Lastly, we use dropout in every stage to address model generalization. Overfitting remains an issue in using deep learning on stereo vision due to lack of large datasets for training. Incidentally, the interconnection between layers and the \textit{Bottleneck Layers} of \textit{DenseNet} also has built-in self-regularizing effect. \textit{Bottleneck Layer} is discussed in the next section.

\section{DenseMapNet}

\begin{figure*}
  \includegraphics[scale=0.31]{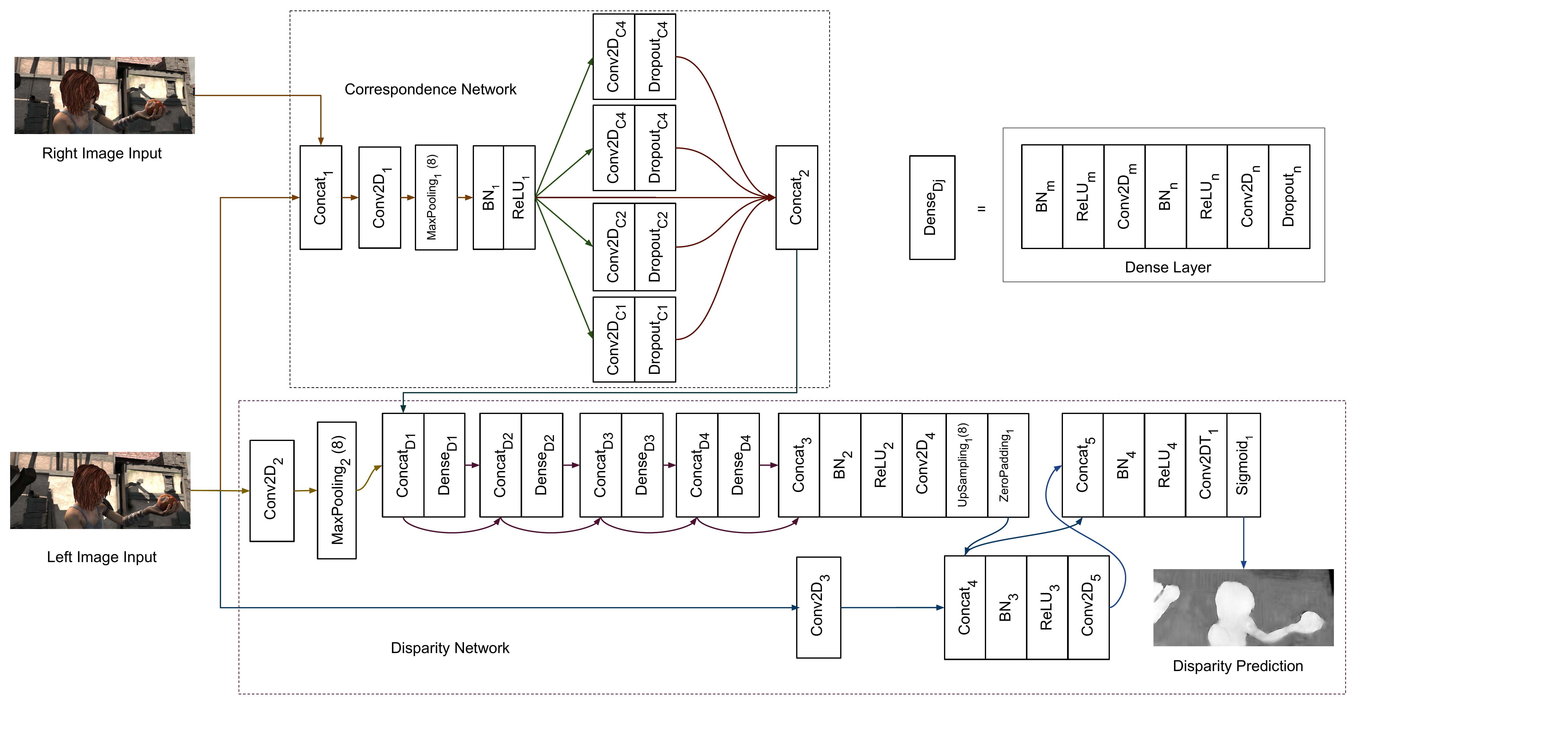}
  \caption{Model architecture of the proposed \textit{DenseMapNet}. The \textit{Correspondence Network} learns how to estimate stereo matching between left and right images. The \textit{Disparity Network} applies the correspondence on the left image. The detail of each \textit{Dense Layer} is also shown.}
  \label{fig:densemapnet}
\end{figure*}

\begin{table}[t]
\caption{Layer description of \textit{DenseMapNet}. $D_i$ is depth (also the number of channels, $C$ if the input is an image). The kernel size is shown as argument in 2D CNN. Dropout rate is 0.2. BN is Batch Normalization with momentum 0.99. \textit{ZeroPadding} is needed only if the result of \textit{UpSampling} does not exactly match the original image dimensions. }
\label{tab:densemapnet_layer}
\begin{minipage}{\columnwidth}
\begin{center}

\begin{tabular}{|c | c | c | c | c|} 
\hline
\scriptsize{Layer} & \scriptsize{Operation} &  \scriptsize{Input Dim} & \scriptsize{Output Dim}\\
\hline
\hline
\tiny{$Concat_1$}& \tiny{Concatenation}  & \tiny{$H\times{W} \times{(C,C)}$} & \tiny{$H\times{W}\times{2C}$}  \\
\hline
\tiny{$Conv2D_1$} & \tiny{2D CNN (5)}  & \tiny{ $H\times{W}\times{2C}$ } & \tiny{ $H\times{W}\times{32}$ }\\
\hline
\tiny{$MaxPooling_i$} & \tiny{ Max Pooling (8)} & \tiny{$H\times{W}\times{D_i}$ } & \tiny{$\frac{H}{8}\times{\frac{W}{8}}\times{D_i}$}\\
\hline
\tiny{$BN_i$} & \tiny{ Batch Normalization } & \tiny{ $H_i\times{W_i}\times{D_i}$ }& \tiny{ $H_i\times{W_i}\times{D_i}$ } \\
\hline
\tiny{$ReLU_i$} & \tiny{Rectified Linear Unit}  & \tiny{$H_i\times{W_i}\times{D_i}$} & \tiny{$H_i\times{W_i}\times{D_i}$} \\
\hline
\tiny{$Dropout_i$} & \tiny{Dropout (0.2)}  & \tiny{$H_i\times{W_i}\times{D_i}$} & \tiny{$H_i\times{W_i}\times{D_i}$} \\
\hline
\tiny{$Conv2D_{Ci}$ }& \tiny{2D CNN (5)} & \tiny{$\frac{H}{8}\times{\frac{W}{8}}\times{32}$}  & \tiny{$\frac{H}{8}\times{\frac{W}{8}}\times{32}$}  \\
\hline
\tiny{$Concat_2$} & \tiny{Concatenation} & \tiny{$\frac{H}{8}\times{\frac{W}{8}}\times{(5\times{32})}$ } & \tiny{$\frac{H}{8}\times{\frac{W}{8}}\times{160}$}  \\
\hline
\tiny{$Conv2D_2$} & \tiny{2D CNN (5)}  & \tiny{$H\times{W}\times{C}$} & \tiny{$H\times{W}\times{16}$} \\
\hline
\tiny{$Concat_{D1}$} & \tiny{Concatenation} & \tiny{$\frac{H}{8}\times{\frac{W}{8}}\times{(16,160)}$}  & \tiny{$\frac{H}{8}\times{\frac{W}{8}}\times{176}$}  \\
\hline
\tiny{$Concat_{D2}$} & \tiny{Concatenation}  & \tiny{$\frac{H}{8}\times{\frac{W}{8}}\times{(16,176)}$}  & \tiny{$\frac{H}{8}\times{\frac{W}{8}}\times{192}$}  \\
\hline
\tiny{$Concat_{D3}$} & \tiny{Concatenation} & \tiny{$\frac{H}{8}\times{\frac{W}{8}}\times{(16,192)}$ } & \tiny{$\frac{H}{8}\times{\frac{W}{8}}\times{208}$}  \\
\hline
\tiny{$Concat_{D4}$} & \tiny{Concatenation}  & \tiny{$\frac{H}{8}\times{\frac{W}{8}}\times{(16,208)}$}  & \tiny{$\frac{H}{8}\times{\frac{W}{8}}\times{224}$}  \\
\hline
\tiny{$Conv2D_m$} & \tiny{2D CNN (1)}  & \tiny{$\frac{H}{8}\times{\frac{W}{8}}\times{D_i}$}  & \tiny{$\frac{H}{8}\times{\frac{W}{8}}\times{64}$}  \\
\hline
\tiny{$Conv2D_n$} & \tiny{2D CNN (5)}  & \tiny{$\frac{H}{8}\times{\frac{W}{8}}\times{64}$}  & \tiny{$\frac{H}{8}\times{\frac{W}{8}}\times{16}$}  \\
\hline
\tiny{$Concat_{3}$} & \tiny{Concatenation} & \tiny{$\frac{H}{8}\times{\frac{W}{8}}\times{(16,224)}$}  & \tiny{$\frac{H}{8}\times{\frac{W}{8}}\times{240}$}  \\
\hline
\tiny{$Conv2D_4$} & \tiny{2D CNN (1)}  & \tiny{$\frac{H}{8}\times{\frac{W}{8}}\times{240}$}  & \tiny{$\frac{H}{8}\times{\frac{W}{8}}\times{32}$}  \\
\hline
\tiny{$Conv2D_3$} & \tiny{2D CNN (5)} & \tiny{$H\times{W}\times{C}$} & \tiny{$H\times{W}\times{1}$} \\
\hline
\tiny{$UpSampling_1$} & \tiny{Up Sampling (8)}  & \tiny{$\frac{H}{8}\times{\frac{W}{8}}\times{32}$}  & \tiny{${H}\times{W}\times{32}$}  \\
\hline
\tiny{$Concat_{4}$} & \tiny{Concatenation}  & \tiny{${H}\times{W}\times{(32,1)}$}  & \tiny{${H}\times{W}\times{33}$}  \\
\hline
\tiny{$Conv2D_5$} & \tiny{2D CNN}  & \tiny{$H\times{W}\times{33}$} & \tiny{$H\times{W}\times{16}$} \\
\hline
\tiny{$Concat_{5}$} & \tiny{Concatenation}  & \tiny{${H}\times{W}\times{(33,16)}$}  & \tiny{${H}\times{W}\times{49}$}  \\
\hline
\tiny{$Conv2DT_1$} & \tiny{2D Transposed CNN (9)} & \tiny{$H\times{W}\times{49}$} & \tiny{$H\times{W}\times{1}$} \\
\hline
\tiny{$Sigmoid{_1}$} & \tiny{Sigmoid}  & \tiny{$H\times{W} \times{1}$} & \tiny{$H\times{W}\times{1}$} \\
\hline
\end{tabular}
\end{center}
\end{minipage}
\end{table}

\begin{figure}
\includegraphics[scale=0.5]{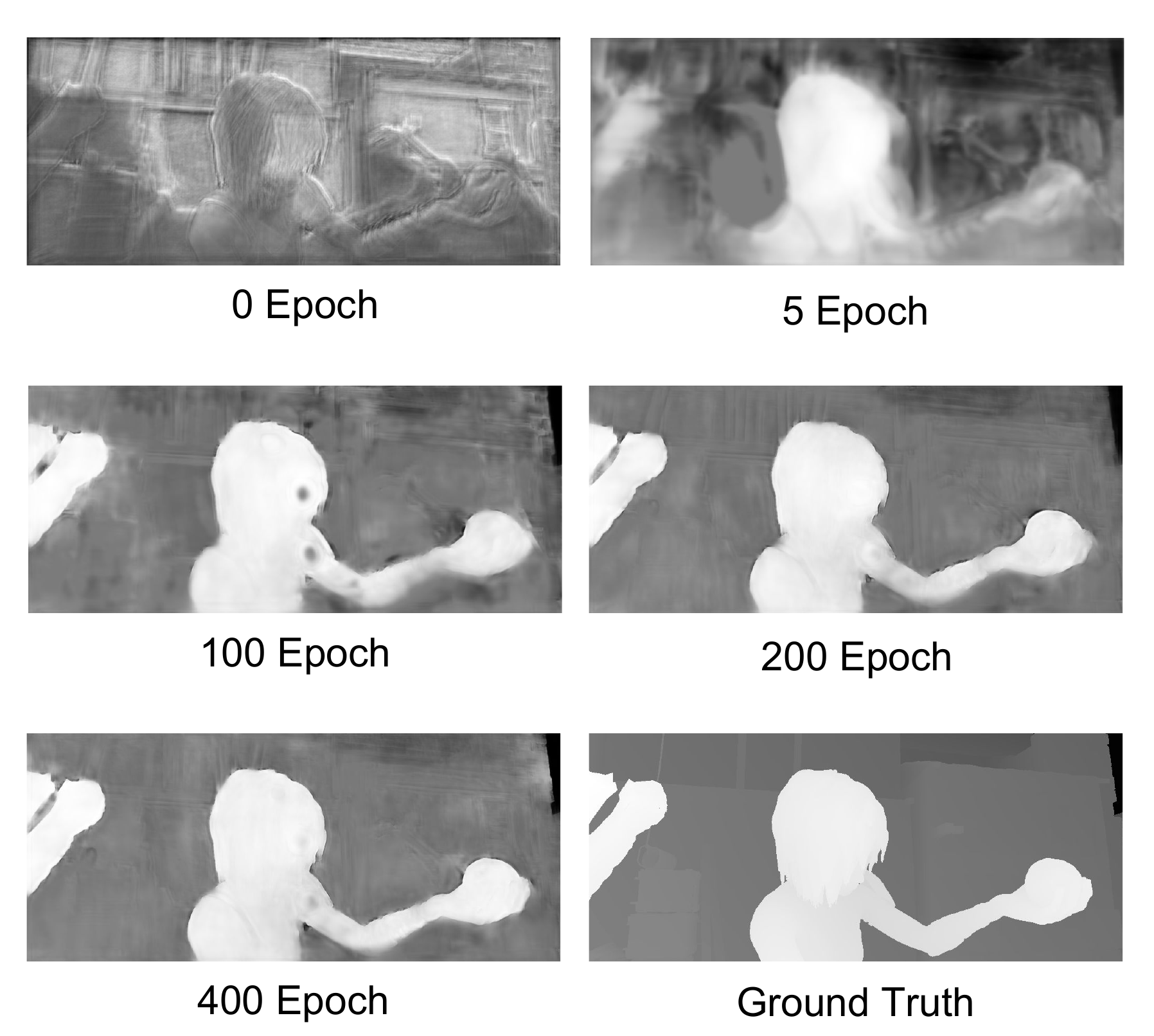}
\caption{In \textit{DenseMapNet}, from the onset of training, the network applies the disparity on the reference (left) image. }
\label{fig:trainimages}
\end{figure}
To generate the correspondence map, the algorithm imposes that we choose a reference image, like the left image. Assuming rectified stereo images,  for each pixel on the left image, we look for the same pixel on the right image by searching on the corresponding row. If there is no occlusion, we will find the match. Otherwise, we  approximate its location. The row column offset is called the disparity.  Given a calibrated camera, we can determine the corresponding depth using Equation~\ref{eq:1}.

For each pixel disparity, we can generate a disparity map using disparity as the measure of intensity or brightness. The intensity is assigned to every pixel in the reference image. Hence, for disparity images, the brighter the object, the closer it is to the camera coordinate system origin.

Using the description of the algorithm for disparity map estimation, we designed \textit{DenseMapNet} as shown in Figure~\ref{fig:densemapnet}. The detailed description of each layer is shown in Table~\ref{tab:densemapnet_layer}. \textit{DenseMapNet} has 18 CNN layers and has two networks to mimic the algorithm for disparity estimation: 1) \textit{Correspondence Network} and 2) \textit{Disparity Network}. The idea is for the \textit{Correspondence Network} to learn stereo matching while the \textit{Disparity Network} applies the disparity on the reference image. 

The \textit{Correspondence Network} aims to find pixel correspondences of stereo images. Hence, instead of making the network deep, the \textit{Correspondence Network} is designed to be wide to increase the coverage of the kernel. $Conv2D_{C1}$ to $Conv2D_{C4}$ use $5\times{5}$ kernels but with increasing dilation rate from 1 to 4. In addition, the stereo images feature maps are reduced by 8 in dimensions to further increase the coverage of the kernel.

The \textit{Disparity Network} utilizes the learned representations from the \textit{Correspondence Network} to estimate the amount of disparity to be applied on the reference image. As shown in Figure~\ref{fig:densemapnet}, the \textit{Disparity Network} processes both feature maps from the reference image (left image) and the \textit{Correspondence Network} to lead  \textit{DenseMapNet} in estimating the disparity map. The \textit{Disparity Network} has 13 CNN layers. As shown in Figure~\ref{fig:trainimages}, from the onset of the training to 400 epochs, the prediction is progressing in a stable manner.

The prominent feature of \textit{DenseMapNet} is the \textit{Dense Network}-type of connection wherein the loss function has access to all feature maps down to the input layer. The output of the immediate previous layer and inputs of all previous layers are inputs to the current layer. The loss function's immediate access to all CNN layers prevents gradients from vanishing as they travel down the shallow layers. Parameter sharing makes this type of CNN efficient by significantly reducing the total number of weights to train. Immediate access to weights and parameter sharing make \textit{DenseMapNet} easy to train. 

In \textit{DenseMapNet}, the role of $Concat_i$ is to combine the previous layer outputs and all previous layers inputs to form the new inputs to the current layer. The inputs include the raw stereo images and their feature maps. $Concat_i$ plays an important role in \textit{Dense Network} to ensure connection of the loss function down to the input images.  

Since every CNN layer's feature maps are connected to the succeeding layers, it is easy for the number of inputs to the deeper layers to escalate. In order to avoid the escalation of the number of inputs, we compress the feature maps. In Table~\ref{tab:densemapnet_layer}, all CNN layers with $1\times1$ kernel compress the feature maps into 64 or 32 layers. These CNN layers are similar in purpose to the \textit{Bottleneck} layers of \textit{DenseNet}. Without the compressing layers, \textit{DenseNet}-like networks are computationally heavy.

The key feature extraction is performed by CNN layers with $5\times{5}$ and $9\times{9}$ kernels in \textit{Correspondence Network} and \textit{Disparity Network}. $Dense_{C1}$ to $Dense_{C4}$ have an expanding dilation rate from 1 to 4 similar in \textit{Correspondence Network} to increase the kernel coverage. The four \textit{Dense Layers} have 64 output feature maps in the compressing layers instead of 32 in \textit{Correspondence Network} since these layers carry the combined disparity measure as applied to the reference image.

The transposed CNN, $Conv2DT_{1}$, performs the final prediction which is scaled to $[0.0, 1.0]$ (equivalent to $[0, max\_pixel\_disparity]$) by the $Sigmoid_1$ layer. $sigmoid(x) = \frac{1}{1 + e^{-x}}$ is suitable for the recovery of gradient descent especially when the predicted value is wrongly pushed to very high or very low values.

All CNN layers have 1) Rectified Linear Unit, $ReLU_i$, activation layer to introduce non-linearity \cite{nair2010rectified} and 2) Batch Normalization layer, $BN_i$, to stabilize the training even at higher learning rates \cite{ioffe2015batch}. Although \textit{DenseNet} by design prevents overfitting, we still use $Dropout_i$ in feature extracting CNN layers \cite{srivastava2014dropout}.

Although \textit{DenseMapNet} input/output are full resolution images, GPU memory is limited. Using $MaxPooling_1$, we downsample the input images and their feature maps to allow us to have wide \textit{Correspondence Network} and \textit{Dense Layers}. In the latter part of \textit{Disparity Network}, we return the feature maps to full resolution using $UpSampling_1$ and perform further feature extraction. $Max Pooling$ is used instead of strided convolution for memory and parameter efficiency and for speed. The same reason why $Up Sampling$ is used instead of strided transposed convolution. 

\begin{figure}
\includegraphics[scale=0.5]{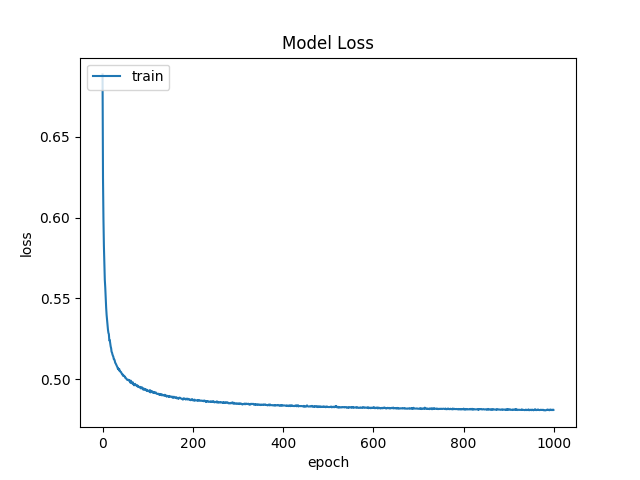}
\caption{Loss function value during training for MPI Sintel dataset.}
\label{fig:loss}
\end{figure}

\begin{table*}
\small
\caption{Benchmark on different datasets. All errors are End-Point-Errors (EPE). Baseline data are from \cite{mayer2016large}.}
\label{tab:results}
\begin{center}
\begin{tabular}{|c | c |c | c | c | c | c | c | c|} 
\hline
Method & Sintel & Driving & FlyingThings3D & Monkaa &  KITTI 2015 & Parameters & Speed & GPU\\
\hline
\hline
DispNet & 5.38 & 15.62 & \underline{2.02} & 5.99 & \underline{2.19} & 38.4M & 16.67Hz & NVIDIA Titan X \\
\hline
SGM & 19.62 & 40.19 & 8.70 & 20.16 & 7.21 &- & 0.91Hz & NVIDIA Titan X \\
\hline
MC-CNN-fast &  11.94 & 19.58 & 4.09 & 6.71 & - & 0.6M & 1.25Hz & NVIDIA Titan X \\
\hline
DenseMapNet & \underline{4.41} & \underline{6.56} & 5.07 & \underline{4.45} & 2.52 & \underline{0.29M} & \underline{$>$30Hz} & NVIDIA GTX 1080Ti \\
\hline
\end{tabular}
\end{center}
\end{table*}

\section{Results and Discussion}
We implemented \textit{DenseMapNet} on Keras \cite{chollet2015keras} with Tensorflow \cite{tensorflow2015-whitepaper} backend. We used NVIDIA GTX 1080Ti GPU for training and testing. The total number of parameters is 290k (0.29M) only. To minimize the binary cross entropy loss between the ground truth and the output of the sigmoid function, \textit{RMSprop} optimizer \cite{hinton2012rmsprop} with learning rate of 1e-3 and decay rate of 1e-6 is used. We also tried MSE as loss function. However, since the error per pixel range $[0.0,1.0]$ is small, it is difficult for the parameters to converge. Our batch size is 4 given the limited memory of the GPU. Figure~\ref{fig:loss} shows the loss function value starts to stabilize at around 500 epochs.

\subsection{Dataset}
We used five publicly available datasets to train and evaluate the performance of \textit{DenseMapNet}:
\begin{enumerate}
  \item \textit{MPI Sintel} \cite{butler2012naturalistic}  has 1064 synthesized stereo images and ground truth data for disparity. \textit{Sintel} is derived from open-source 3D animated short film Sintel. The dataset has 23 different scenes. The stereo images are RGB while the disparity is grayscale. Both have resolution of ${1024}\times{436}$ pixels and 8-bit per channel.
  \item \textit{Driving} \cite{mayer2016large} tries to mimic scenes from KITTI 2012 and 2015 datasets \cite{Menze2015CVPR,Geiger2012CVPR}. The \textit{Driving} dataset contains over 4000  synthesized  image pairs. Each input image is 8-bit RGB and has resolution of $960\times{540}$ pixels. Maximum disparity value is 355 pixels.
  \item \textit{Monkaa} \cite{mayer2016large} is from an open-source animated movie rendered in Blender. There are over 8,000  synthesized stereo images in this dataset. \textit{Monkaa} has the same image specifications as \textit{Driving}. Maximum disparity value is 10,500 pixels.  
  \item \textit{FlyingThings3D} \cite{mayer2016large} has over 25,000  synthesized stereo images of everyday objects that are flying around. We used the test subset of \textit{FlyingThings3D} which has over 4,000  synthesized  image pairs. \textit{FlyingThings3D} has the same image specifications as \textit{Driving}. Maximum disparity value is 6,772 pixels. 
  \item \textit{KITTI 2015} \cite{Menze2015CVPR} has 200 grayscale stereo images of real road scenes with ${1241}\times{376}$ pixels resolution obtained from a stereo rig mounted on a vehicle. LIDAR is used to established the ground truth depth maps. Hence, the disparity maps are sparse. We cropped the stereo images and disparity maps to ${1224}\times{200}$ pixels or the regions with most LIDAR measurements available. Maximum disparity is $43887/256$ pixels.
\end{enumerate}
  We randomly shuffled each entire dataset and set aside 90\% for training and 10\% for testing. We removed samples with unrealistic disparity values (greater than the image width) from \textit{Monkaa} and \textit{FlyingThings3D}.
\subsection{Results}
Table~\ref{tab:results} shows the benchmark test results of \textit{DenseMapNet} compared to the state-of-the-art \textit{DispNet}. Also shown are the results for SGM and MC-CNN-fast to established the baselines. The biggest advantage of our network is speed and size. With only 0.29M parameters, it is 0.7\% in size of \textit{DispNet} and 48\% of MC-CNN-fast. With small size, \textit{DenseMapNet} processes a pair of images for disparity estimation at 30Hz or faster. For \textit{MPI Sintel}, the speed is 35Hz while for \textit{KITTI 2015} which is nearly half of full resolution is 66Hz. Even with a slower GPU, our network is nearly twice as fast as the fastest CNN-based disparity estimation method to date. With regards to accuracy, \textit{DenseMapNet} performance is better except for \textit{FlyingThings3D} and \textit{KITTI 2015}. It has the highest accuracy improvement on \textit{Driving} dataset at 6.56 EPE, 58\% reduction in error compared to the highest accuracy established. We believe that the EPE on \textit{KITTI 2015} is statistically comparable to the state-of-the-art.

\begin{figure}
\includegraphics[scale=0.3]{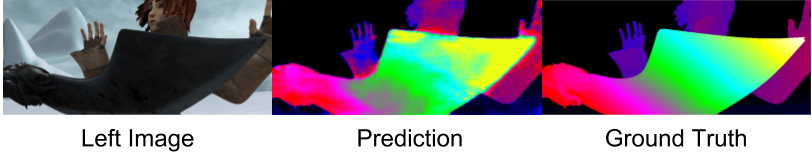}
\caption{Sample predicted disparity maps of \textit{DenseMapNet}. Also shown are left input image and ground truth.}
\label{fig:mpi}
\end{figure}

\begin{figure}
\includegraphics[scale=0.29]{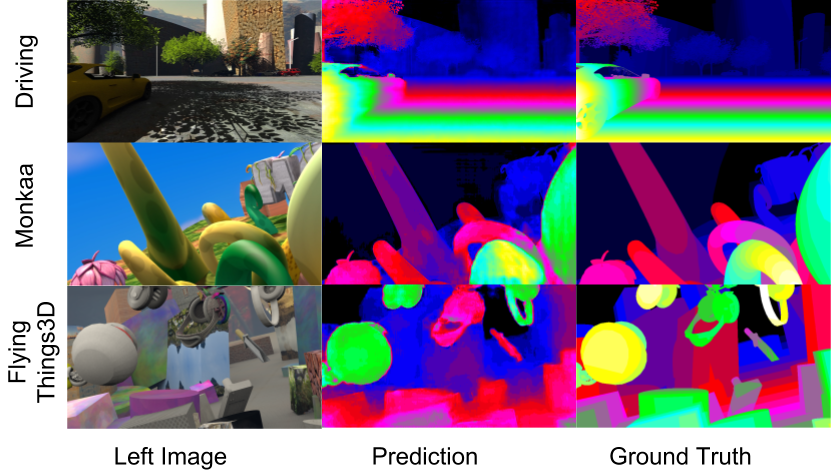}
\caption{Sample predicted disparity maps of \textit{DenseMapNet} for \textit{Driving}, \textit{Monkaa}, and \textit{FlyingThings3D} datasets.}
\label{fig:driving-monkaa-flying}
\end{figure}

\begin{figure}
\includegraphics[scale=0.4]{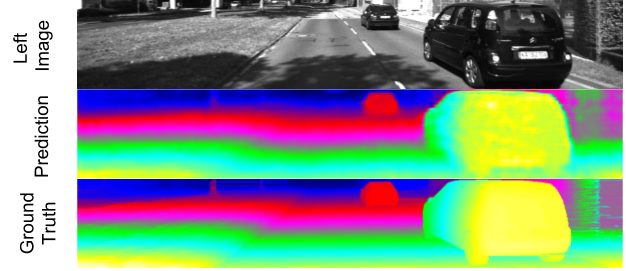}
\caption{Sample predicted disparity map of \textit{DenseMapNet} for \textit{KITTI 2015} dataset.}
\label{fig:kitti2015}
\end{figure}

Figures~\ref{fig:mpi} to \ref{fig:kitti2015} demonstrate sample predicted disparity maps of \textit{DenseMapNet}. Also shown are ground truth disparity maps and reference images. We use the color map of KITTI 2012 \cite{Geiger2012CVPR} to show contrast on disparity maps. The accompanying video and its longer version at Youtube, \href{https://youtu.be/NBL-hFQRh4k}{https://youtu.be/NBL-hFQRh4k}, demonstrate fast disparity estimation of \textit{DenseMapNet} on sequence of images.

\section{CONCLUSIONS}
\textit{DenseMapNet} demonstrates that it is possible to arrive at a compact and fast CNN model architecture by taking advantage of semantics and interconnection in feature maps. Our proposed model is suitable for computation and memory constrained machines like drones and other autonomous robots. Codes and other results of \textit{DenseMapNet} can be found in our project repository: \href{https://github.com/roatienza/densemapnet}{https://github.com/roatienza/densemapnet}.





{\small
\bibliographystyle{IEEEtran}
\bibliography{egbib}
}

\end{document}